\documentclass[twocolumn]{article}
\usepackage{graphicx} 
\usepackage{todonotes}
\usepackage{hyperref}
\usepackage{subcaption}
\usepackage{natbib}
\usepackage{enumitem}
\usepackage{amsmath,mathabx}
\usepackage[top=22mm, bottom=22mm, left=22mm, right=22mm]{geometry}

\title{We are not able to identify AI-generated images}
\author{Adrien Pav\~ao}
\date{}

\begin{document}

\maketitle

\section*{Abstract}

AI-generated images are now pervasive online, yet many people believe they can easily tell them apart from real photographs. We test this assumption through an interactive web experiment where participants classify 20 images as real or AI-generated. Our dataset contains 120 difficult cases: real images sampled from CC12M, and carefully curated AI-generated counterparts produced with MidJourney. In total, 165 users completed 233 sessions. Their average accuracy was 54\%, only slightly above random guessing, with limited improvement across repeated attempts. Response times averaged 7.3 seconds, and some images were consistently more deceptive than others. These results indicate that, even on relatively simple portrait images, humans struggle to reliably detect AI-generated content. As synthetic media continues to improve, human judgment alone is becoming insufficient for distinguishing real from artificial data. These findings highlight the need for greater awareness and ethical guidelines as AI-generated media becomes increasingly indistinguishable from reality.

\section{Introduction}

AI-generated content is becoming a dominant part of the online ecosystem. Images, text, music, and even videos produced by machine-learning models now circulate alongside authentic human-made content, often without being clearly identified as such.
We make the hypothesis that most people overestimate their capacity to reliably spot AI-generated images. This confidence rise from a perception bias: individuals mainly remember the cases where they guessed correctly, while the times they were fooled often pass unnoticed.
To measure how well humans can truly distinguish real images from AI-generated ones, we collected an images dataset and built an interactive web experiment, accessible at the following URL:

\begin{center}
\url{https://adrienpavao.com/RealOrAI}
\end{center}

Participants are shown a sequence of 20 images and must classify each one as either real photo or AI-generated. Their performance allows us to quantify human accuracy, identify the most deceptive images, and analyze how decision time and prior exposure affect the results.

{\bf Recent studies} was performed on similar protocols, on text data \citep{sota2025}, on images \citep{sota2025b} and mixing audio, images, and text \citep{sota2023}. The results suggest performance only slightly above random guessing.


\section{Data and Methods}

\begin{figure}[ht]
    \centering
    \begin{subfigure}{0.15\textwidth}
        \centering
        \includegraphics[width=\linewidth]{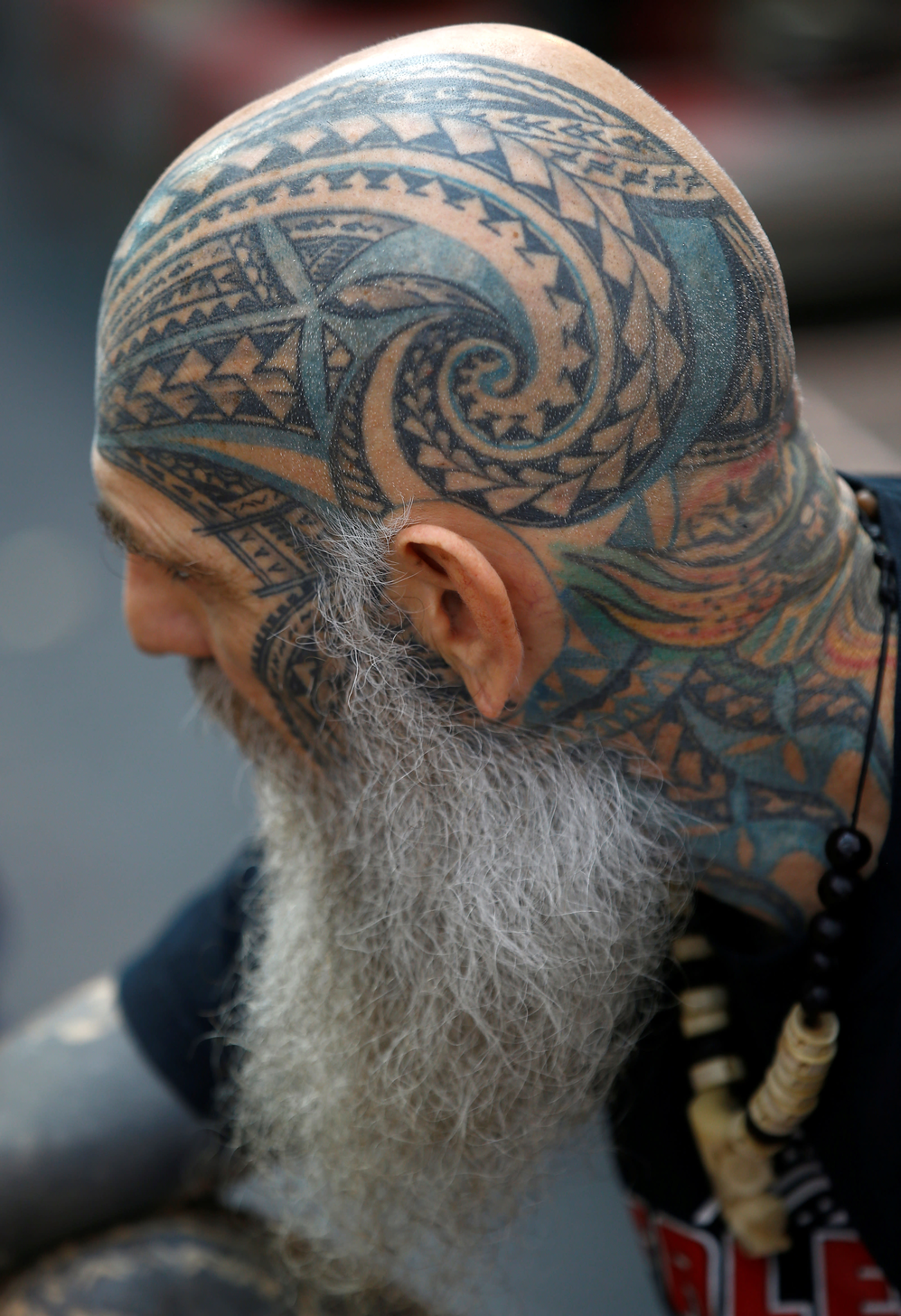}
    \end{subfigure}
    \hfill
    \begin{subfigure}{0.15\textwidth}
        \centering
        \includegraphics[width=\linewidth]{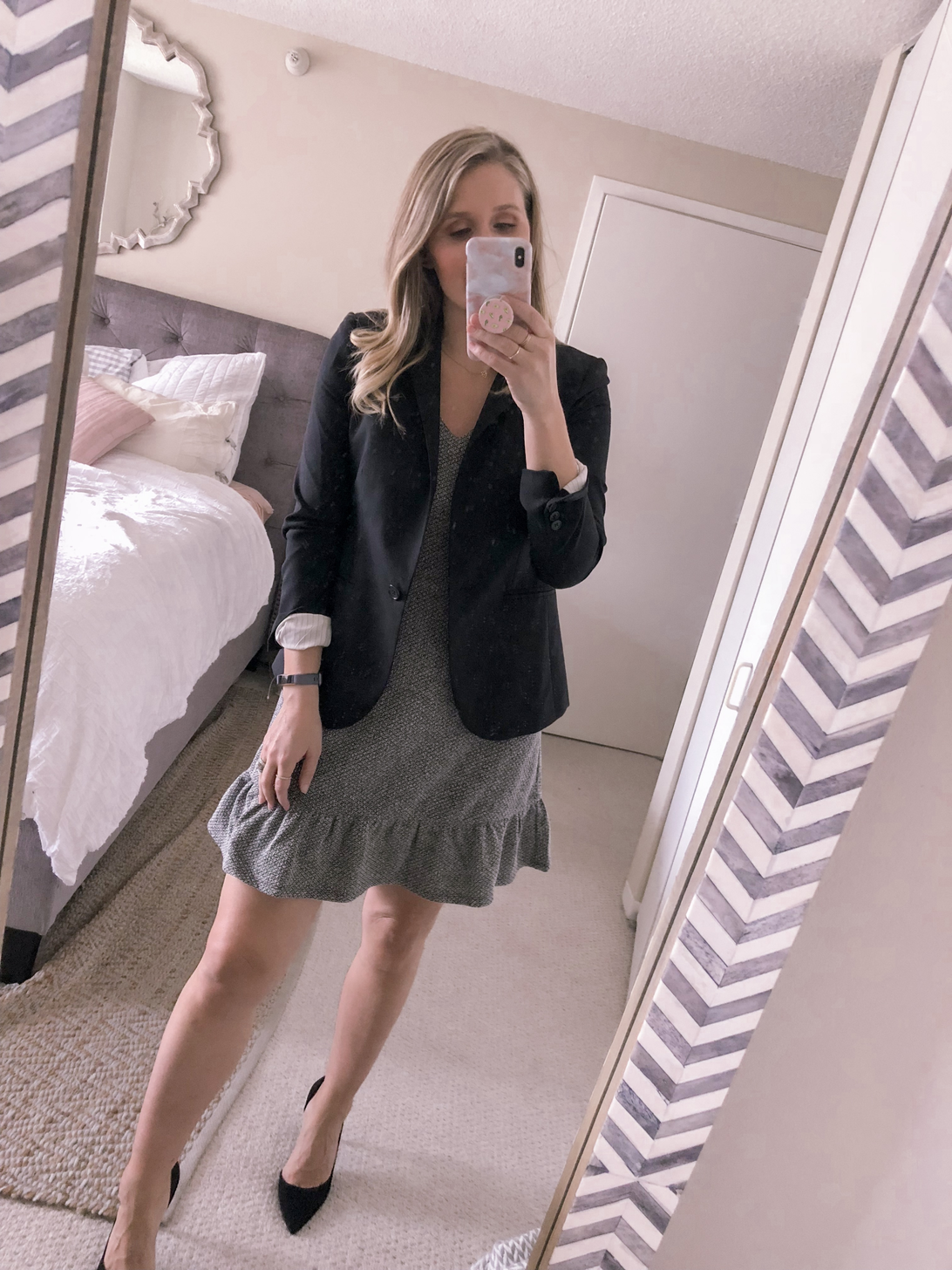}
    \end{subfigure}
    \hfill
    \begin{subfigure}{0.15\textwidth}
        \centering
        \includegraphics[width=\linewidth]{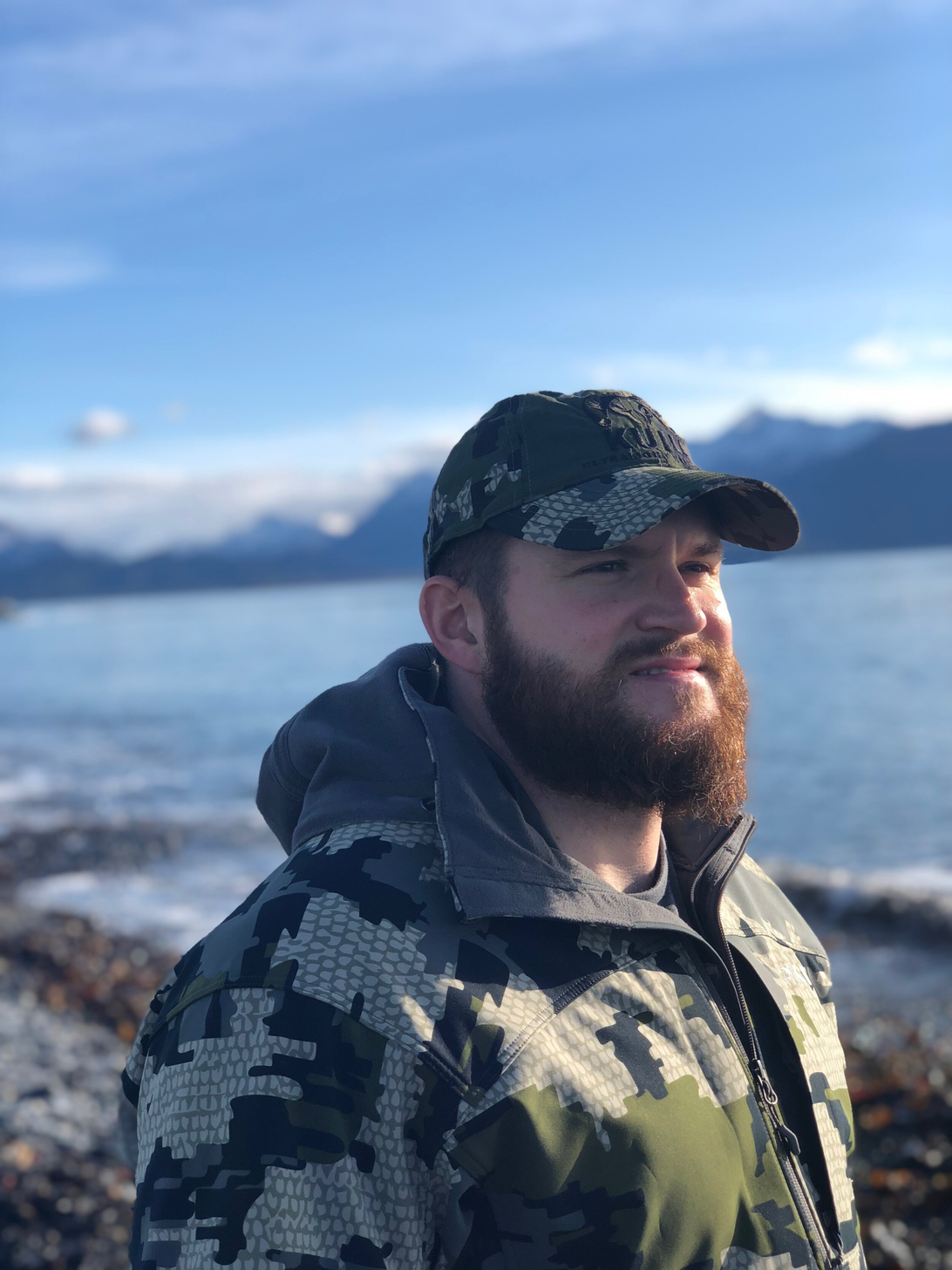}
    \end{subfigure}
    \vspace{0.4cm}
    \begin{subfigure}{0.15\textwidth}
        \centering
        \includegraphics[width=\linewidth]{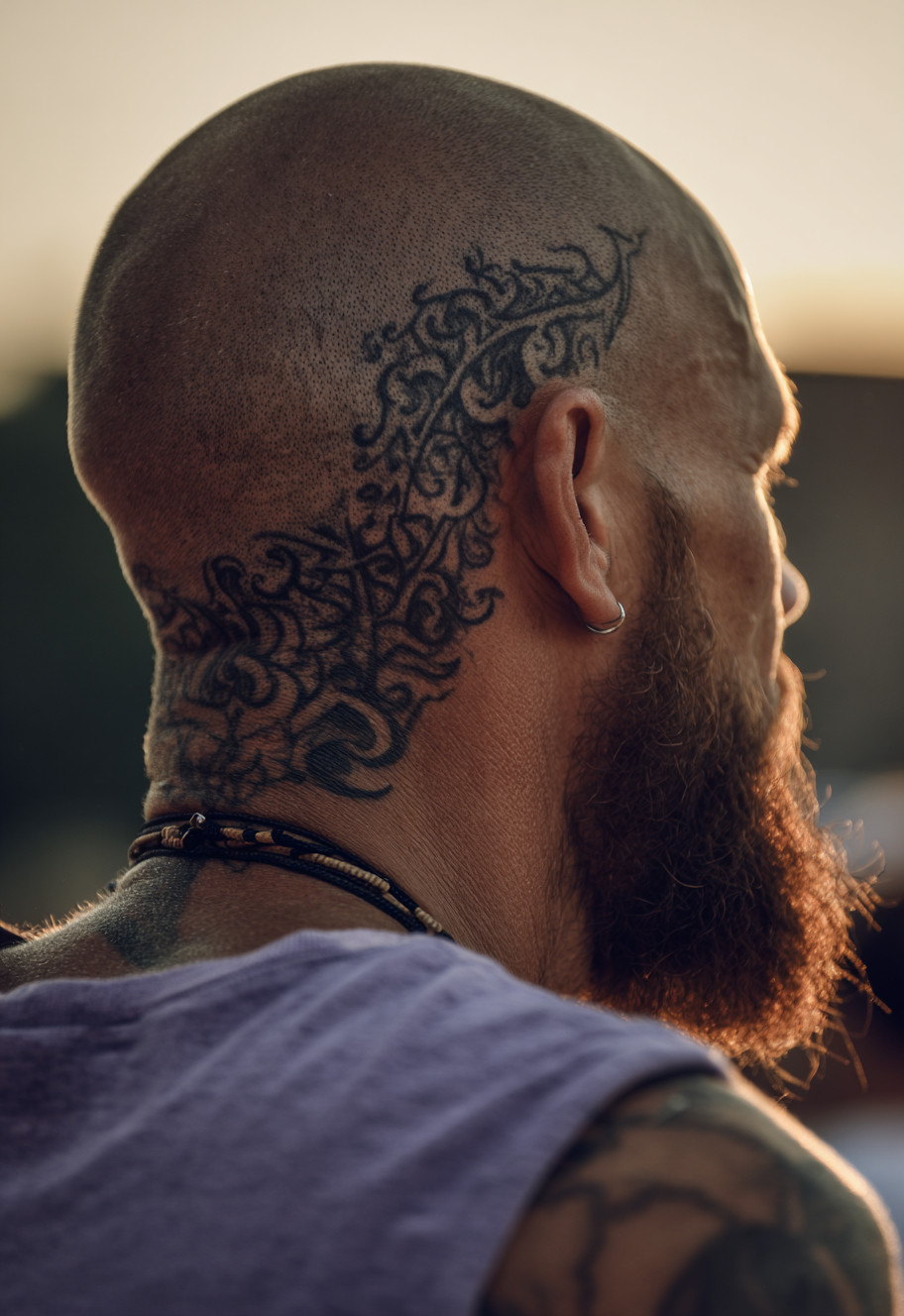}
    \end{subfigure}
    \hfill
    \begin{subfigure}{0.15\textwidth}
        \centering
        \includegraphics[width=\linewidth]{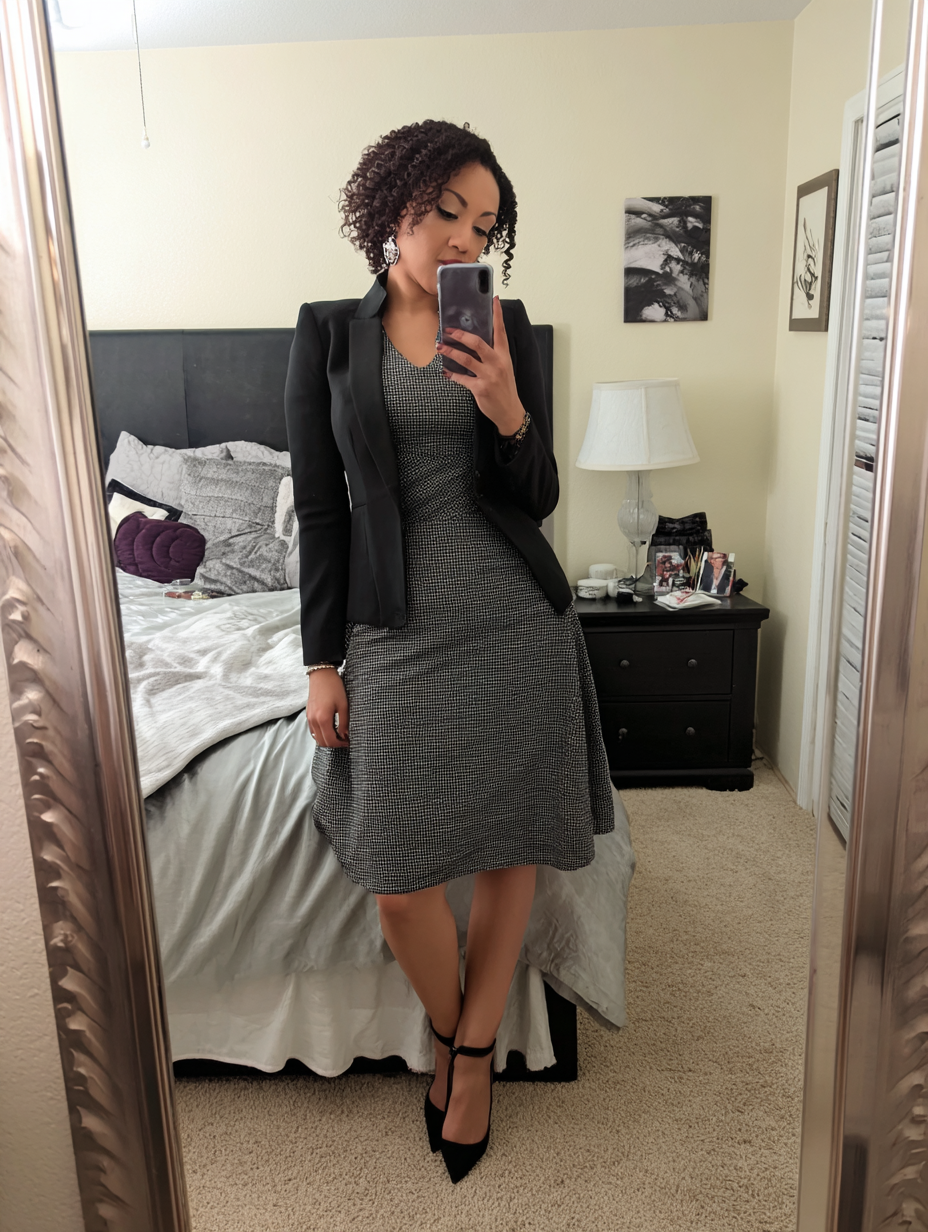}
    \end{subfigure}
    \hfill
    \begin{subfigure}{0.15\textwidth}
        \centering
        \includegraphics[width=\linewidth]{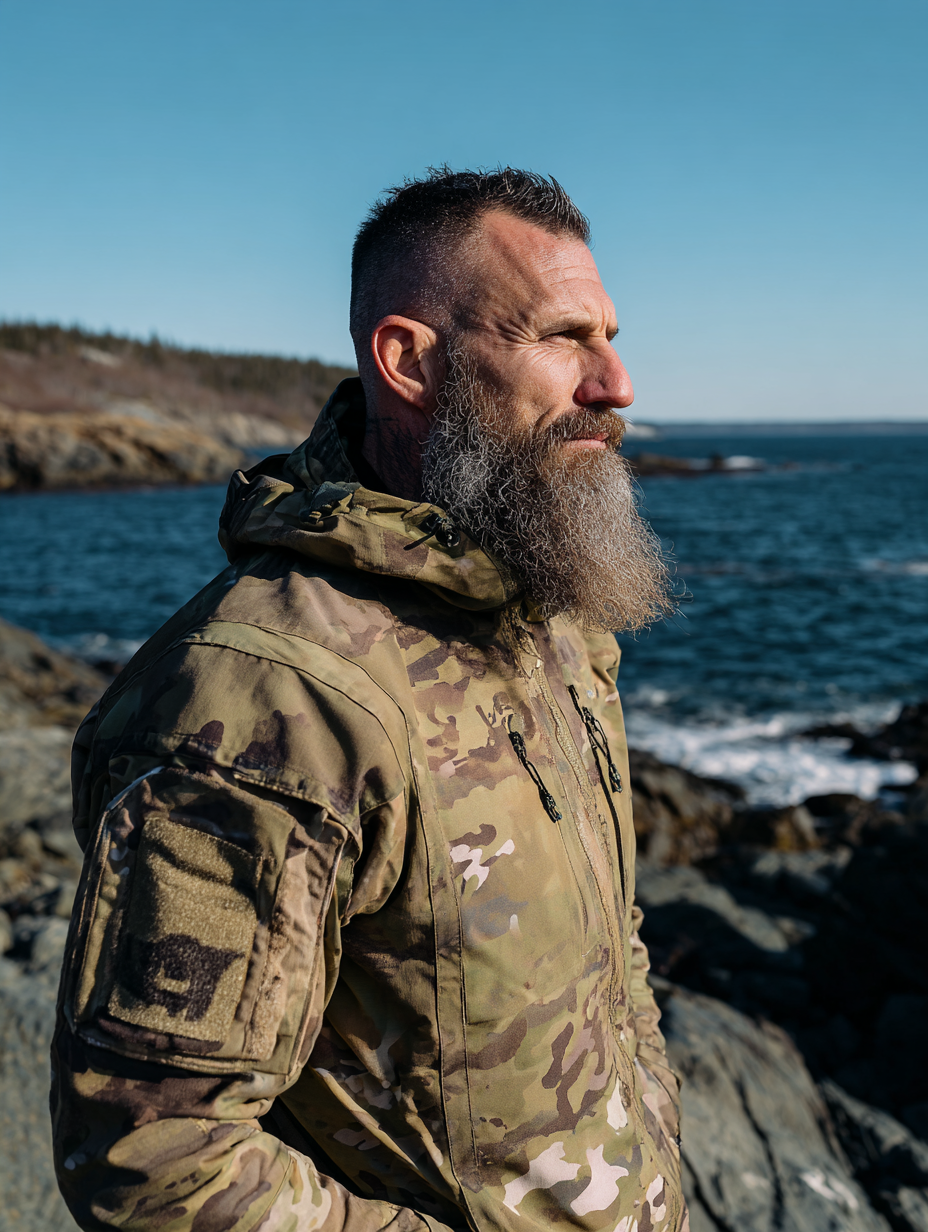}
    \end{subfigure}
    \caption{Examples from the dataset: three real images (top row) and three AI-generated images (bottom row).}
    \label{fig:examples}
\end{figure}

Our dataset is intentionally small (120 images) but designed to be difficult. This first version acts as a proof of concept before extending the project to larger image sets, as well as other modalities such as music or video. The dataset is balanced, with 60 images depicting women and 60 depicting men. Examples of real and AI-generated images are shown in Figure \ref{fig:examples}.

{\bf Real images} were sampled from the CC12M dataset \citep{cc12m}.  
We selected random entries whose text description started by either ``a man'' or ``a woman''. Obvious celebrities were removed to avoid easy recognition. Images were then filtered for quality and resized to a consistent resolution.

{\bf Fake images} were generated using MidJourney v7 \citep{midjourney}. For each selected real image, we extracted its original CC12M text description and image dimension and used it as a prompt seed. We ran each prompt several times with similar parameters (for example: \texttt{--stylize 50}, \texttt{--v 7}) and manually selected the most realistic results. This curation step was essential, as many generated candidates still looked artificial or stylized. We acknowledge that this introduces a selection bias: the AI-generated images in our dataset are not a random sample, but the most convincing ones among several generations. We consider this bias acceptable, and even desirable, for the purpose of this study. Our goal is to evaluate human performance on the hardest cases, not on obviously synthetic images. Moreover, in real online settings, people who share AI-generated portraits or photographs typically select the most realistic outputs rather than posting the first attempt. Our curation process therefore reflects common user behavior.

Here are a few examples of prompts used:

\begin{itemize}[label=-]
    \item ``A woman stands in front of a restaurant door and smiles. --ar 3:4 --style raw --v 7 --stylize 50’’
    \item ``A man walks through an airport with high ceilings and large windows. --ar 3:2 --style raw --v 7 --stylize 50’
    \item ``A woman walks the runway in a striking red and black dress with a wavy pattern, showcasing her profile. --ar 1667:2500 --style raw --v 7 --stylize 50’’
\end{itemize}

Using actual descriptions from the real data as prompts was done in order to ensure a balanced distribution in scenery, postures and clothes between real and fake data. We acknowledge that this design introduces a form of bias, where participants exposed to similar pictures may infer that they are not part of the same set (either real or fake), thus improving their guessing score.

{\bf The test} consists of 20 images, randomly sampled without replacement. They are shown one by one, and the user must answer ``Real'' or ``AI''. The system records each answer and the response time. Results are saved only once the session is complete. Users can then restart the game; unique users are identified so repeated sessions can be linked and avoided as a source of bias in the analysis.

The full labeled dataset, as well as the experiment results, can be found here:

\begin{center}
\url{https://github.com/didayolo/RealOrAI-dataset}
\end{center}


\section{Results}

\begin{figure}
    \centering
    \includegraphics[width=0.9\linewidth]{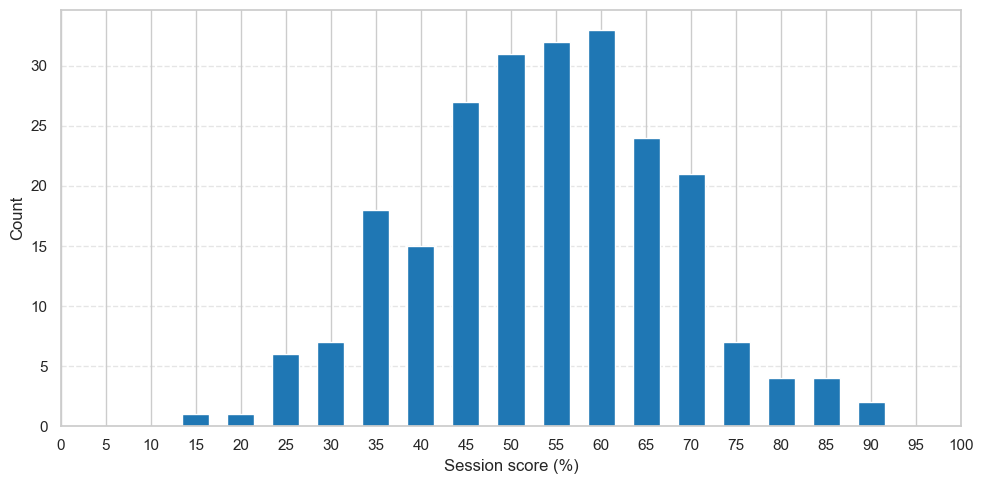}
    \caption{Scores distribution}
    \label{fig:scores}
\end{figure}

In total, {\bf 165 users} performed {\bf 233 sessions} of 20 images. In average, they obtained an {\bf accuracy of 53.76\%} (SD = 14\%, SE = 1\%), showing very little capacity to distinguish real and AI-generated images. The scores seem to follow a normal distribution, as shown by Figure \ref{fig:scores}. We observe a discrepancy in score between the two classes, as the average accuracy on real pictures is slightly better that on artificial ones, $55.76\%$ and $51.78\%$ respectively.

\begin{figure}
    \centering
    \includegraphics[width=0.9\linewidth]{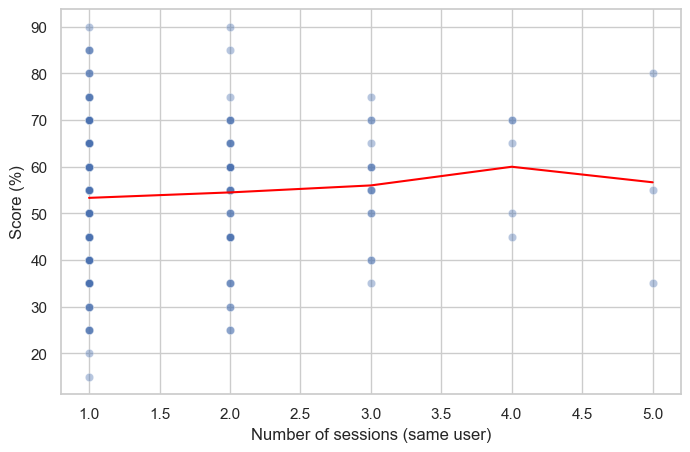}
    \caption{Users progress, scatter and tendency. Average score obtained when engaging in sessions.}
    \label{fig:sessions}
\end{figure}

Users engaging in multiple sessions only obtained a slight improvements in their scores, as shown in Figure \ref{fig:sessions}. The average accuracy on first session is 53.33\%. This is not surprising, as there is a total of 120 images, so we do not except many data repetition on a small number of sessions.

\begin{figure}
    \centering
    \includegraphics[width=0.9\linewidth]{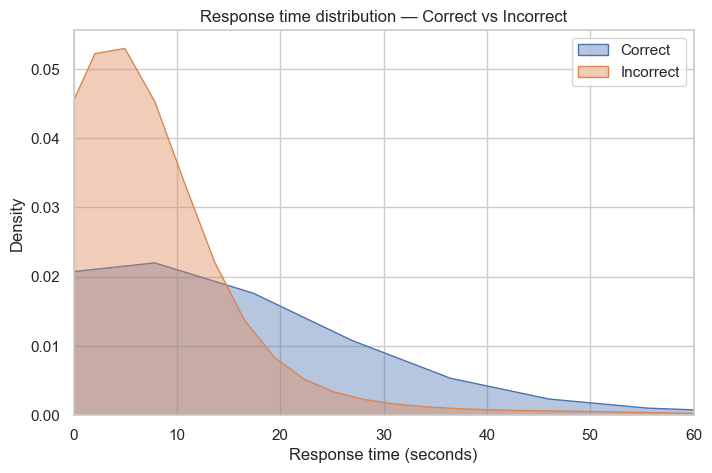}
    \caption{Response time density for correct and incorrect answers.}
    \label{fig:time-density}
\end{figure}

\begin{figure}
    \centering
    \includegraphics[width=0.9\linewidth]{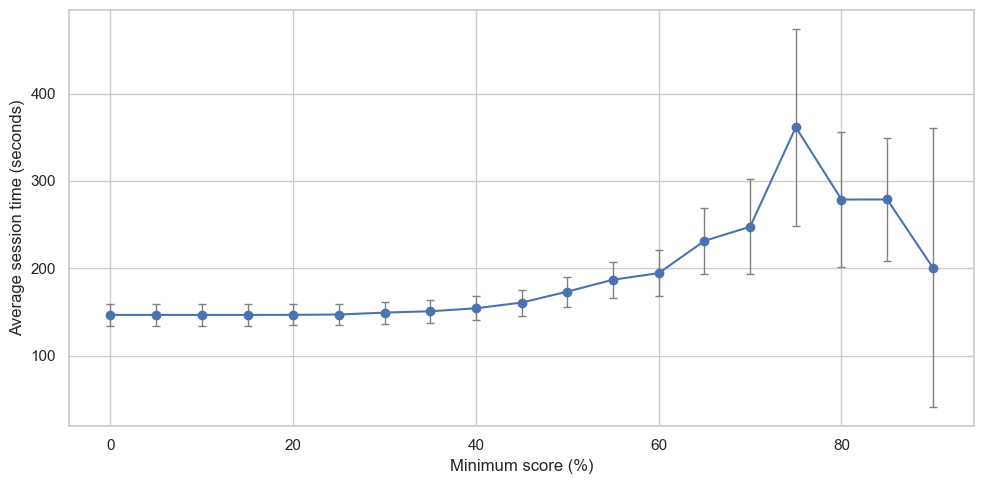}
    \caption{Average time for sessions with score $\geq x$. Error bars show the standard error on the measurements.}
    \label{fig:time-score}
\end{figure}

On average, participants responded in {\bf 7.33 seconds} per image.
The response-time distributions indicate that fast decisions yield accuracy close to random, whereas accuracy improves steadily with longer viewing times. The curves for correct and incorrect responses intersect around 15 seconds, suggesting that deliberate inspection is required to exceed chance performance (Figure \ref{fig:time-density}).
Participants with higher final scores also tend to complete sessions more slowly, with a marked increase in average session duration above roughly 60\%, presented in Figure \ref{fig:time-score}. This pattern supports the idea that accurate discrimination relies on slower, more careful analysis rather than rapid intuition.

\begin{figure*}[t]
    \centering
    \includegraphics[width=\textwidth]{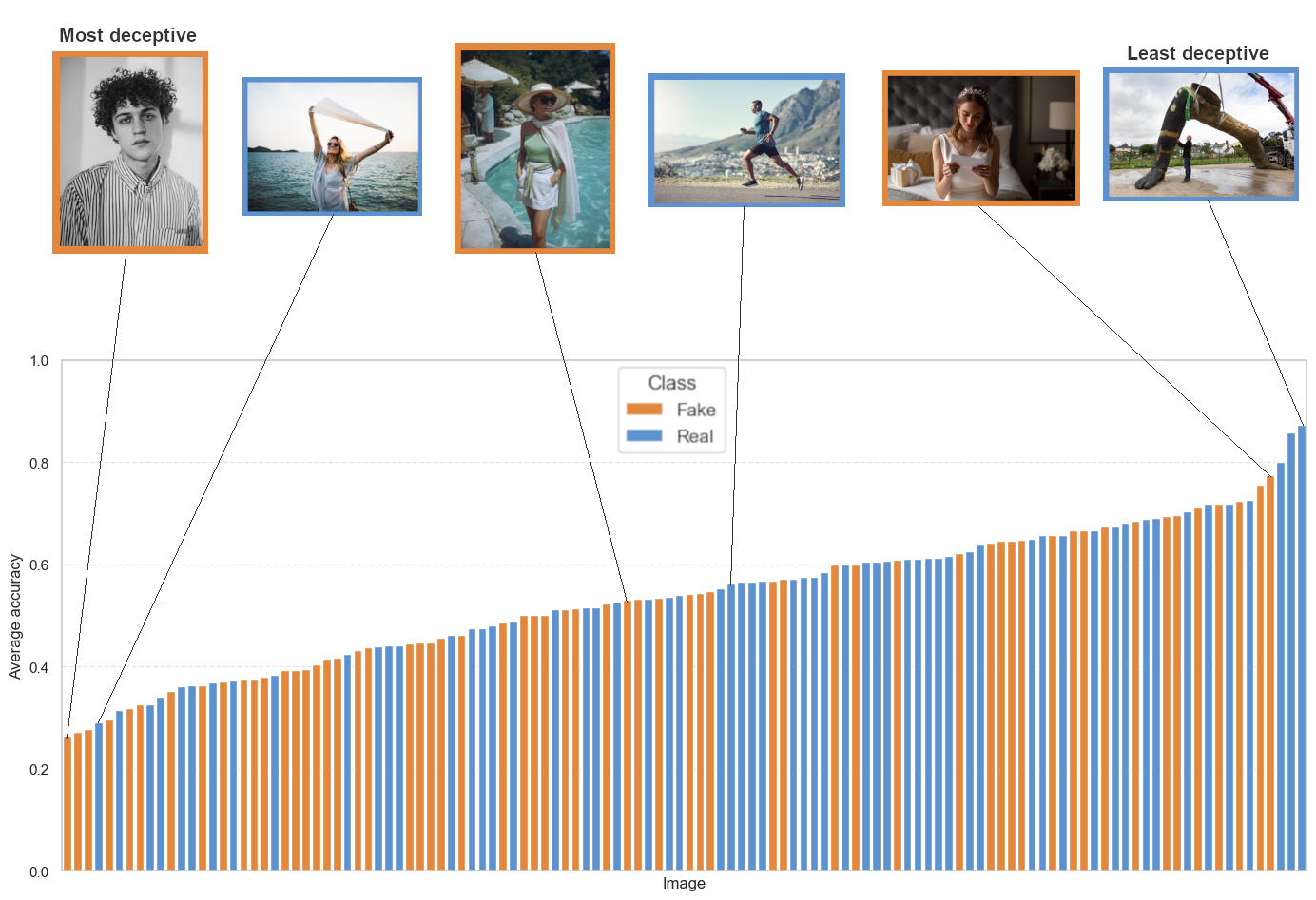}
    \caption{Performance obtained by the users on each individual images. 6 example images are shown, sorted out by most to least deceptive.}
    \label{fig:deception}
\end{figure*}

It is also interesting to note that users' performance vary across images, ranging from 26\% to 87\% accuracy. Each image average accuracy, or difficulty, is presented in Figure \ref{fig:deception}, as well as example of the least and most deceptive fake and real images.

\section{Discussion and conclusion}

Our results confirm that humans struggle to distinguish real images from AI-generated ones, even under conditions that should favor careful inspection. With an average accuracy of 54\%, participants performed only slightly above random guessing. The narrow margin above chance, together with the limited improvement across repeated sessions, suggests that the visual cues distinguishing real from AI-generated portraits are either extremely subtle or effectively absent for most users.

Response-time analyses add an important layer to this finding. Participants required more than ten seconds on average to reach accuracy levels meaningfully above chance, and the deliberate inspection implied by the experiment clearly improved performance. On social media, in messaging apps, or while browsing online, images are typically evaluated in a fraction of a second, often without any explicit suspicion that they might be synthetic. In such ``scrolling-speed'' conditions, the accuracy observed in our experiment would likely be even lower. This gap between experimental focus and real-world attention helps explain why realistic AI-generated images can be so effective at passing unnoticed.

Scores also varied across images, indicating that some examples were consistently more deceptive than others. This variation suggests that not all AI-generated portraits are equally difficult, and that specific combinations of lighting, framing, or subject pose may increase or reduce detectability. A more systematic study of these factors could provide insight into what makes an image inherently deceptive.

The composition of our participant pool may introduce a bias. The experiment circulated mainly through personal networks, social media, and professional connections, leading to a sample likely skewed toward individuals familiar with AI or digital media. If anything, this bias should favor higher accuracy; the difficulty of the task for a more representative population is therefore likely underestimated.

Several limitations point to directions for future work. First, the dataset remains relatively small, and expanding it would improve the robustness of the analysis. Second, the study focuses solely on still images; extending the protocol to video, audio, or multimodal content would better reflect the range of synthetic media encountered online. Finally, replacing binary answers with confidence scores could reveal whether users have reliable uncertainty estimates even when their classifications are incorrect.

Taken together, these findings show that human judgment alone is insufficient for reliably distinguishing real from AI-generated imagery, even in simple portrait scenarios. As synthetic media continues to improve, awareness, labeling practices, and technical detection tools will be essential components of a broader strategy to maintain trust in visual information.

\bibliographystyle{plainnat}
\bibliography{ref}

@inproceedings{cc12m,
  title = {{Conceptual 12M}: Pushing Web-Scale Image-Text Pre-Training To Recognize Long-Tail Visual Concepts},
  author = {Changpinyo, Soravit and Sharma, Piyush and Ding, Nan and Soricut, Radu},
  booktitle = {CVPR},
  year = {2021},
}

@misc{midjourney,
  title        = {MidJourney},
  author       = {MidJourney},
  howpublished = {\url{https://www.midjourney.com}},
  year         = {2022}
}

@article{sota2025,
title = {Do humans identify AI-generated text better than machines? Evidence based on excerpts from German theses},
journal = {International Review of Economics Education},
volume = {49},
pages = {100321},
year = {2025},
issn = {1477-3880},
doi = {https://doi.org/10.1016/j.iree.2025.100321},
url = {https://www.sciencedirect.com/science/article/pii/S1477388025000131},
author = {Alexandra Fiedler and Jörg Döpke},
}

@misc{sota2025b,
      title={How good are humans at detecting AI-generated images? Learnings from an experiment}, 
      author={Thomas Roca and Anthony Cintron Roman and Jehú Torres Vega and Marcelo Duarte and Pengce Wang and Kevin White and Amit Misra and Juan Lavista Ferres},
      year={2025},
      eprint={2507.18640},
      archivePrefix={arXiv},
      primaryClass={cs.HC},
      url={https://arxiv.org/abs/2507.18640}, 
}

@misc{sota2023,
      title={A Representative Study on Human Detection of Artificially Generated Media Across Countries}, 
      author={Joel Frank and Franziska Herbert and Jonas Ricker and Lea Schönherr and Thorsten Eisenhofer and Asja Fischer and Markus Dürmuth and Thorsten Holz},
      year={2023},
      eprint={2312.05976},
      archivePrefix={arXiv},
      primaryClass={cs.CR},
      url={https://arxiv.org/abs/2312.05976}, 
}

\end{document}